\documentclass[11pt,a4paper]{article}
\usepackage{times,latexsym}
\usepackage{url}
\usepackage[T1]{fontenc}

\usepackage[acceptedWithA]{tacl2021v1}

\usepackage{xspace,mfirstuc,tabulary}

\newif\iftaclinstructions
\taclinstructionsfalse 
\iftaclinstructions
\renewcommand{\confidential}{}
\renewcommand{\anonsubtext}{(No author info supplied here, for consistency with
TACL-submission anonymization requirements)}
\newcommand{\instr}
\fi

\iftaclpubformat 

\else

\fi

\usepackage{dsfont}
\usepackage{bm}
\usepackage{bbm}
\usepackage{graphicx}
\usepackage{color}
\usepackage{multicol}
\usepackage{multirow}
\usepackage{wrapfig,lipsum,booktabs}
\usepackage[ruled,vlined,linesnumbered]{algorithm2e}
\usepackage{pgfplots}
\pgfplotsset{compat=1.12} 
\usepackage{filecontents}
\usepackage{xcolor}
\usepackage{tikz}
\usepackage{xspace}
\usepackage{makecell}
\usepackage{soul}
\usepackage{amsmath,amsfonts,amssymb}
\usepackage{subcaption}
\usetikzlibrary{calc}
\usepgfplotslibrary{groupplots}
\usetikzlibrary{angles,quotes} 
\usetikzlibrary{shapes,arrows}
\usetikzlibrary{backgrounds}
\usetikzlibrary{matrix}
\usetikzlibrary{patterns}
\usepackage{tikz-3dplot}
\usepackage{hyperref}
\usepackage{cleveref}
\usepackage{paralist}
\usepackage{cancel}
\usepackage{xspace}
\usepackage{todonotes}
\usepackage{tabu}
\usepackage{rotating}
\usepackage{etoolbox}
\usepackage{adjustbox}
\usepackage{enumerate}
\usepackage{enumitem}
\setitemize{itemsep=0pt,topsep=0pt,parsep=0pt,partopsep=0pt}
\setenumerate{itemsep=0pt,topsep=0pt,parsep=0pt,partopsep=0pt}
\usepackage{cancel}
\usepackage{lipsum}
\usepackage{listings,lstautogobble}
\usepackage{fancyvrb}
\usepackage{fvextra}
\usepackage{caption}
\usepackage{CJKutf8}
\usepackage{colortbl}

\newcommand{\model}[1]{\textsc{#1}\xspace}

\newcommand{\ours}{\model{TransAgents}}
\newcommand{\gptfour}{\model{gpt-4-turbo}}
\newcommand{\dbleu}{$d$-\model{BLEU}}
\newcommand{\gemba}{\model{Gemba-DA}}
\newcommand{\gpto}{\model{gpt-4o}}
\newcommand{\gptomini}{\model{gpt-4o-mini}}
\newcommand{\revised}[1]{\textcolor{black}{#1}\xspace}

\usepackage{amsmath,amsfonts,bm}

\def\eqref#1{equation~\ref{#1}}

\def\1{\bm{1}}

\def\rmA{{\mathbf{A}}}

\def\rmC{{\mathbf{C}}}
\def\rmD{{\mathbf{D}}}

\def\rmF{{\mathbf{F}}}

\def\rmH{{\mathbf{H}}}
\def\rmI{{\mathbf{I}}}
\def\rmJ{{\mathbf{J}}}

\def\rmM{{\mathbf{M}}}

\def\rmP{{\mathbf{P}}}
\def\rmQ{{\mathbf{Q}}}
\def\rmR{{\mathbf{R}}}
\def\rmS{{\mathbf{S}}}

\DeclareMathAlphabet{\mathsfit}{\encodingdefault}{\sfdefault}{m}{sl}
\SetMathAlphabet{\mathsfit}{bold}{\encodingdefault}{\sfdefault}{bx}{n}

\title{(Perhaps) Beyond Human Translation: Harnessing Multi-Agent Collaboration for Translating Ultra-Long Literary Texts}

\author{
Minghao Wu\textsuperscript{1}\quad Jiahao Xu\textsuperscript{2}\quad Yulin Yuan\textsuperscript{3}\quad Gholamreza Haffari\textsuperscript{1}\quad\\
{\bf Longyue Wang\textsuperscript{4$\dagger$} \quad Weihua Luo\textsuperscript{4}\quad Kaifu Zhang\textsuperscript{4}}\\
\textsuperscript{1}Monash University \quad \textsuperscript{2}Nanyang Technological University 
\\ \textsuperscript{3}University of Macau \quad \textsuperscript{4}Alibaba International Digital Commerce\\
\textsuperscript{1}\texttt{\{minghao.wu, gholamreza.haffari\}@monash.edu}\\
\textsuperscript{2}\texttt{Jiahao004@e.ntu.edu.sg} \quad \textsuperscript{3}\texttt{yulinyuan@um.edu.mo}\\
\textsuperscript{4}\texttt{\{wanglongyue.wly, weihua.luowh, kaifu.zkf\}@alibaba-inc.com}
}

\date{}

\begin{document}

\renewcommand{\tableautorefname}{Table}
\renewcommand{\sectionautorefname}{Section}
\renewcommand{\subsectionautorefname}{Section}
\renewcommand{\subsubsectionautorefname}{Section}
\renewcommand{\figureautorefname}{Figure}
\renewcommand{\equationautorefname}{Equation}
\renewcommand{\algorithmautorefname}{Algorithm}
\newcommand{\linenoautorefname}{Line}

\maketitle

\begingroup
\renewcommand\thefootnote{}\footnotetext{
\textsuperscript{$\dagger$}~Longyue Wang is the corresponding author.
}
\endgroup

\begin{abstract}

\revised{
Literary translation remains one of the most challenging frontiers in machine translation due to the complexity of capturing figurative language, cultural nuances, and unique stylistic elements. In this work, we introduce \ours, a novel multi-agent framework that simulates the roles and collaborative practices of a human translation company, including a CEO, Senior Editor, Junior Editor, Translator, Localization Specialist, and Proofreader. The translation process is divided into two stages: a preparation stage where the team is assembled and comprehensive translation guidelines are drafted, and an execution stage that involves sequential translation, localization, proofreading, and a final quality check. Furthermore, we propose two innovative evaluation strategies: Monolingual Human Preference (MHP), which evaluates translations based solely on target language quality and cultural appropriateness, and Bilingual LLM Preference (BLP), which leverages large language models like \model{gpt-4} for direct text comparison. Although \ours achieves lower \dbleu scores, due to the limited diversity of references, its translations are significantly better than those of other baselines and are preferred by both human evaluators and LLMs over traditional human references and \model{gpt-4} translations. Our findings highlight the potential of multi-agent collaboration in enhancing translation quality, particularly for longer texts.}\footnote{GitHub Repository: \url{https://github.com/minghao-wu/transagents}}

\end{abstract}

\section{Introduction}

Machine translation (MT) has achieved significant advancements recently, driven by breakthroughs in deep learning and neural networks \citep{cho-etal-2014-learning, DBLP:conf/nips/SutskeverVL14, DBLP:conf/nips/VaswaniSPUJGKP17, gu2019nips, liu-etal-2020-multilingual-denoising, fan2021jmlr}. Despite these technological strides, literary translation remains a challenging area for MT systems. Literary texts, with their complex language, figurative expressions, cultural nuances, and unique stylistic elements, pose significant hurdles that are difficult for machines to overcome \citep{voigt-jurafsky-2012-towards}. This complexity makes literary translation one of the most challenging areas within machine translation, referred to as ``the last frontier of machine translation'' \citep{Jeremy2024}.

Recent research in multi-agent systems, especially those using large language models (LLMs), shows significant promise \citep{DBLP:conf/iclr/YaoZYDSN023, DBLP:journals/corr/abs-2302-01560, DBLP:journals/corr/abs-2304-07590}. These systems harness the collective intelligence of multiple agents, outperforming individual models in dynamic environments that require complex problem-solving and collaboration. 
\revised{
In this work, we leverage LLMs to simulate the various roles in a human translation company and propose \ours, a multi-agent framework designed to mimic the best practices of human translation. The translation process in \ours is divided into two main stages: preparation and execution, each comprising several sub-stages. In the preparation stage, the designated CEO agent selects a Senior Editor based on the specific needs of the client. The Senior Editor then assembles a team from a roster of roles, including Junior Editor, Translator, Localization Specialist, and Proofreader. Together, the Senior Editor and Junior Editor draft a comprehensive translation guideline to ensure consistency and quality throughout the process. During the execution stage, the Translator agent translates the text chapter by chapter, adhering to the guidelines. The Localization Specialist revises the translated text to ensure cultural and contextual alignment with the target audience. Subsequently, the Proofreader reviews the translation to eliminate errors and verify adherence to the guideline. Finally, the Senior Editor conducts a thorough quality check to ensure the final translation meets the highest standards. By dividing the workflow into distinct stages and assigning specialized roles, \ours effectively emulates the structured approach of human translation companies, leveraging the capabilities of LLMs to enhance translation quality and efficiency.
}

Furthermore, evaluating the accuracy and quality of literary translations is challenging due to the subjective nature of literature and the potential imperfections in reference translations \citep{thai-etal-2022-exploring,freitag-etal-2023-results}. To address these challenges, in addition to conventional MT evaluation metrics, we propose two innovative evaluation strategies: \textit{Monolingual Human Preference} (MHP) and \textit{Bilingual LLM Preference} (BLP). Monolingual Human Preference involves human evaluators from the target audience assessing translations without referring to the original text. This approach focuses on fluidity, readability, and cultural appropriateness, simulating the real-world consumption of literature. Bilingual LLM Preference (BLP) leverages advanced LLMs, such as \model{gpt-4}, which are provided with the original texts to facilitate direct comparison. This method aims to utilize the superior translation capabilities of LLMs, mitigating the impact of imperfect reference translations.

\revised{
Our empirical findings reveal that, while \ours achieves the lowest \dbleu scores, it significantly outperforms the state-of-the-art machine translation (MT) method in terms of \gemba. Additionally, both human evaluators and a large language model (LLM) evaluator prefer the translations produced by \ours over human-written references and \model{gpt-4} translations. Our analysis suggests that the significant decline in \dbleu scores is primarily due to the limited diversity of the reference translations. In contrast, the translations generated by \ours exhibit significantly greater diversity compared to other approaches. Furthermore, we demonstrate the effectiveness of our design in terms of agent profiling and collaboration strategies. Interestingly, while \ours excels in translating long texts and significantly outperforms other baselines in this area, it struggles with translating shorter texts effectively.
}

Our contributions are summarized as follows:
\begin{itemize}
    \item We present \ours, a novel multi-agent system for literary translation that emulates the traditional translation publication process. This multi-agent approach effectively addresses the complex nuances inherent in literary works.
    \item We propose two novel evaluation strategies, \textit{Monolingual Human Preference} (MHP) and \textit{Bilingual LLM Preference} (BLP) to assess the quality of translations. These methods address the limitations of traditional machine translation evaluation metrics.
    \item Despite achieving lower \dbleu scores, our empirical results demonstrate that translations produced by \ours are significantly better than other approaches in terms of \gemba and preferred by both human evaluators and language models over human references and \model{GPT-4} translations. Furthermore, we offer a comprehensive analysis of the strengths and weaknesses of \ours.
\end{itemize}


\section{Related Work}

\paragraph{Machine Translation}
Machine translation (MT) has seen significant advancements recently
However, these improvements are mainly at the sentence level. Recent efforts focus on integrating contextual information to enhance translation quality beyond individual sentences \citep{wang-etal-2017-exploiting-cross, wu-etal-2023-document, herold-ney-2023-improving, wu-etal-2024-importance}. Large language models (LLMs) have also shown superior capabilities in MT \citep{DBLP:journals/corr/abs-2309-11674,robinson-etal-2023-chatgpt,wang-etal-2023-document-level,DBLP:journals/corr/abs-2401-06468}. Despite progress, MT performance in the general domain is saturating, shifting interest towards literary translation, which demands accuracy and cultural nuance. 
\revised{
\citet{karpinska-iyyer-2023-large} show that while LLMs effectively leverage document-level context for literary translation, they still make critical errors. 
\citet{wang-etal-2024-benchmarking} evaluate LLMs' ability to translate long texts and propose context extrapolation to improve translation quality. Additionally, recent research explores using LLMs for evaluating literary translations \citep{DBLP:journals/corr/abs-2407-03658,DBLP:journals/corr/abs-2410-18697}.
}
In this work, we introduce a novel multi-agent virtual company for literary translation and propose two evaluation strategies for assessing translation quality.

\paragraph{Multi-Agent Systems}
Intelligent agents are designed to understand their environments, make informed decisions, and respond appropriately \citep{DBLP:journals/ker/WooldridgeJ95}. Compared to single-agent setups, multi-agent systems leverage collaboration among multiple agents based on LLMs to tackle complex problems or simulate real-world environments effectively \citep{DBLP:journals/corr/abs-2402-01680}. Recent studies have shown promising outcomes in areas such as software development \citep{DBLP:journals/corr/abs-2307-07924,DBLP:journals/corr/abs-2308-00352}, multi-robot collaboration \citep{DBLP:journals/corr/abs-2307-04738,DBLP:journals/corr/abs-2307-02485}, evaluation \citep{DBLP:journals/corr/abs-2308-07201,DBLP:journals/corr/abs-2404-18796}, and fact-checking \citep{DBLP:journals/corr/abs-2305-14325}. Additionally, extensive research explores using multiple agents to simulate societal, economic, and gaming environments \citep{DBLP:conf/uist/ParkPCMLB22,DBLP:conf/uist/ParkOCMLB23,DBLP:journals/corr/abs-2309-04658,DBLP:journals/corr/abs-2310-10436,DBLP:journals/corr/abs-2310-08901}. \citet{DBLP:journals/corr/abs-2305-19118} proposes leveraging multi-agent debate for machine translation. However, their approach is limited to the sentence level. 

\revised{
\paragraph{Ours}
In this work, we utilize LLMs to replicate the traditional translation workflow employed by human translation companies, streamlining the process step by step and present a demo \citep{wu-etal-2024-transagents}. Recently, \citet{briakou-etal-2024-translating} introduce a step-by-step approach for translating long texts. Additionally, \citet{DBLP:journals/corr/abs-2410-08143} proposed a multi-agent framework designed to preserve long-term memory in document-level machine translation. These studies are concurrent with our work and share similar goals of enhancing translation quality through structured methodologies.
}

\section{\ours: A Multi-Agent Company for Literary Translation}




\subsection{Company Organization}
\label{sec:overview}

To simulate the entire book translation process, \ours involves a diverse range of roles besides the CEO, including senior editors, junior editors, translators, localization specialists, and proofreaders, each with distinct responsibilities: 
\begin{itemize}
    \item \textbf{Senior Editors}: Senior editors are responsible for overseeing the content production process. Their primary duties encompass setting editorial standards, guiding other team members, and making decisions regarding publication schedules and content direction.
    \item \textbf{Junior Editors}: Junior editors work under senior editors, managing day-to-day editorial workflows, editing content, and assisting in content planning, and handling communications with various roles within the team.
    \item \textbf{Translators}: Translators convert written material from one language to another, preserving the original text's tone, style, and context.
    \item \textbf{Localization Specialists}: Localization specialists adapt content for specific regions or markets, translating and adjusting cultural references to resonate with local audiences.
    \item \textbf{Proofreaders}: Proofreaders perform final checks for grammar, spelling, punctuation, and other possible errors.
\end{itemize}

\begin{figure}[t]
    \centering
    \footnotesize
    \begin{Verbatim}[frame=single, fontsize=\tiny, breaklines=true, breakanywhere=true, commandchars=\\\{\}]
Name: Sofia Chang
Languages: English, Mandarin, Spanish, French
Nationality: Canadian
Gender: Female
Age: 47
Education: Ph.D. in Comparative Literature
Personality: meticulous, introverted, perfectionist, critical, thoughtful
Hobbies: gardening, chess, watercolor painting
Rate per word: 0.12
Years of working: 22
Profession: Senior Editor
Role prompt: You are Sofia Chang, a highly esteemed Senior Editor [TRUNCATED]
    \end{Verbatim}
    \caption{An example profile of \textbf{Senior Editor}.}
    \label{fig:profile_example}
\end{figure}

\paragraph{Agentization} To enhance the realism and efficacy of our translation process simulation, we use \gptfour to generate 30 diverse virtual agent profiles for each role. As shown in \autoref{fig:profile_example}, these profiles encompass a wide range of attributes beyond language skills, including gender, nationality, rate per word, education, experience, and more. This detailed approach enriches the simulation's authenticity and reflects the complexity and diversity of real-world translation settings, thereby supporting and inspiring future research.


\subsection{Translation Workflow}
\label{sec:workflow}

\begin{figure}[t]
    \centering
    \includegraphics[scale=0.35]{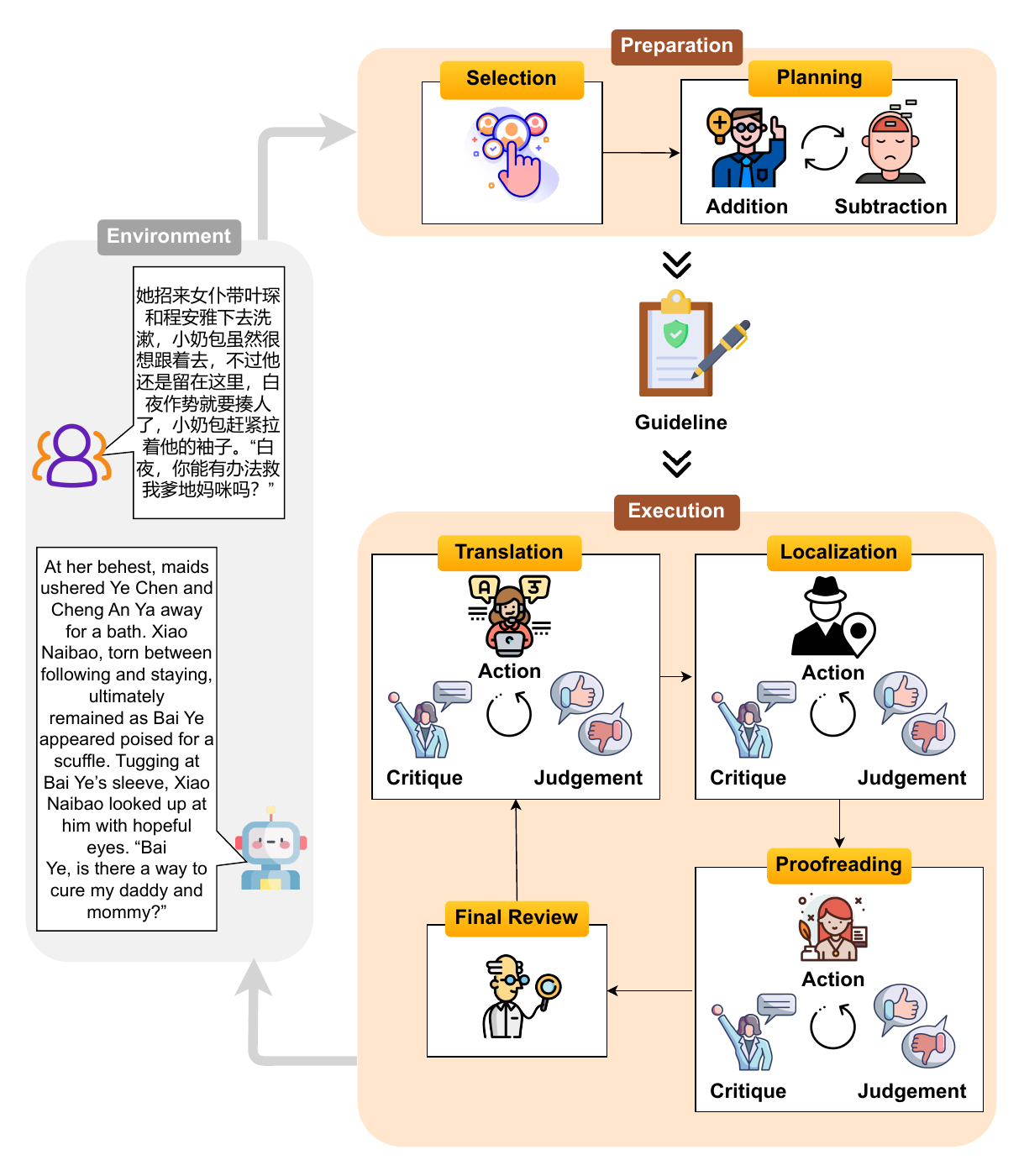}
    \caption{The workflow of \ours.}
    \label{fig:workflow}
\end{figure}

\revised{As visualized in \autoref{fig:workflow},} we introduce the book translation workflow in our company \ours, including two main stages: preparation (\autoref{sec:prep}) and execution (\autoref{sec:exec}), in this section. 

\subsubsection{Preparation}
\label{sec:prep}

\paragraph{Project Member Selection}
In our company, we create 30 agent profiles for each role, each with a unique role assignment prompt (see \autoref{fig:profile_example}). These prompts help assign specific roles to agents before dialogues begin. Initially, the CEO selects a Senior Editor for the book translation project, considering the client's requirements and the candidates' qualifications. Once chosen, the Senior Editor collaborates with the CEO to form the project team, taking into account the candidates' skills and backgrounds. Additionally, we employ a \textit{self-reflection} strategy \citep{DBLP:conf/iclr/YaoZYDSN023, shinn2023reflexion, DBLP:journals/corr/abs-2307-07924} by using a ``ghost agent'' to prompt the CEO to reconsider their decision, as they sometimes struggle to select a Senior Editor with the necessary language skills.

\begin{algorithm}[t]
\SetKwInOut{Input}{Input}
\SetKwInOut{Output}{Output}
\Input{\revised{$\rmC$, the context; $\rmI$, the instruction; $\rmM$, the maximum number of iterations; $m$, the index of current iteration; $\rmA$, the Addition agent; $\rmS$, the Subtraction agent; $\rmF$, the feedback from the Subtraction agent;  $\rmR'$, the temporary response;}}
\Output{\revised{$\rmR$, the final response;}}
$\rmH \leftarrow [\rmC; \rmI]$ \quad $\triangleright$ Initialize the conversation;\\
$\rmR \leftarrow \emptyset$ \quad $\triangleright$ Initialize the response;\\
$m \leftarrow 0$ \quad $\triangleright$ Current round;\\
\While{$m < \rmM$}{
$m \leftarrow m+1$;\\
$\rmR' \leftarrow \rmA(\rmH)$ \quad $\triangleright$ Generate detailed response;\\
$\rmF \leftarrow \rmS(\rmH, \rmR')$ \quad $\triangleright$ Review and remove redundant information;\\
$\rmH \leftarrow \rmH + [\rmR';\rmF]$ \quad $\triangleright$ Append $\rmR'$ and $\rmF$ to the conversation history $\rmH$;\\
\If{$\rmR = \rmR'$}{\label{line:add_by_sub_exit}
Break \quad $\triangleright$ Stop iterating as no further revisions are needed;
}
$\rmR \leftarrow \rmR'$;
}
Return the final response $\rmR$;

\caption{Addition-by-Subtraction Collaboration}
\label{alg:add_by_sub}
\end{algorithm}

\revised{
\paragraph{Addition-by-Subtraction Collaboration}
In our framework, we introduce the \textit{Addition-by-Subtraction Collaboration} between two agents. Unlike the debate-style strategy, which involves multiple agents proposing answers and a third-party agent concluding, our method uses only two agents. One is the \textit{Addition agent}, who extracts extracting as much relevant information as possible, and the other is the \textit{Subtraction agent}, who reviews and eliminates redundant details, providing feedback to the Addition agent.  We present the details of our collaboration strategy in \autoref{alg:add_by_sub}. The Addition agent $\rmA$ generates an initial response with comprehensive content. The Subtraction agent $\rmS$ then reviews this response, removes redundancies, and the agents iterate this process until no further revisions are necessary. The iteration process can terminate before reaching the maximum number of iterations if the ``stop'' condition is met. This approach is referred to as the ``early exit'' mechanism.
}
\revised{Note that the maximum number of iterations used in \autoref{alg:add_by_sub} is 2.}


\begin{CJK*}{UTF8}{gbsn}

\begin{figure}[t]
    \centering
    \footnotesize
    \begin{Verbatim}[frame=single, fontsize=\tiny, breaklines=true, breakanywhere=true, commandchars=\\\{\}]
# Translation Guidelines

## Glossary
罗德: Rhode
虚空之龙: Void Dragon
星月佣兵团: Star Moon Mercenary Corps
[TRUNCATED]

## Book Summary
The book centers on Rhode Alante, initially a Summoner Swordsman in the game 'Dragon Soul Continent,' [TRUNCATED]

## Tone
The tone of the book is adventurous and immersive with elements of fantasy and suspense. [TRUNCATED]

## Style
The book is a gripping blend of fantasy and litRPG, characterized by its immersive world-building, dynamic combat scenes, and a clear progression system. [TRUNCATED]

## Target Audience
The target audience for this book includes young adults and adults who enjoy fantasy and adventure genres, particularly those who are fans of MMORPG [TRUNCATED]

# Chapter Text
序章 传奇落幕
乌云笼罩着天空，昏暗无光的地面上四周都是一片狼籍。
[TRUNCATED]

# Instruction
Translate the chapter text from Chinese into English. Ensure that your translation closely adheres to the provided translation guidelines [TRUNCATED]

    \end{Verbatim}
    \caption{An example prompt for the Translator, including the translation guidelines, the chapter text in the source language, and the instruction.}
    \label{fig:guideline}
\end{figure}

\end{CJK*}

\paragraph{Long-Term Memory Management}
One of the principal challenges in literary translation is to maintain the coherence, cohesion, and consistency throughout the translated text. In this work, we classify memory into two types: : \textit{short-term memory}, which pertains to the context within the current conversation, and \textit{long-term memory}, which relates to the context spanning the entire book. Recent efforts allow LLMs for effectively processing the sequence containing several thousands of tokens \citep{DBLP:journals/corr/abs-2309-16039,DBLP:journals/corr/abs-2312-00752,DBLP:journals/corr/abs-2404-07839}, but they remain inadequate for handling the broader context required for entire books \citep{DBLP:journals/corr/abs-2406-16264}. Addressing this limitation, we employ multi-agent collaboration to develop a comprehensive set of translation guidelines, guiding all the agents involved in the translation process. In \ours, there are five components in the translation guidelines: the glossary, the book summary, the tone, the style, and the target audience. We design different strategies to process them:
\begin{itemize}
    \item \textbf{Glossary and Book Summary}: The glossary of the book compiles essential terms from the source language and provides their corresponding translations in the target language, while the book summary provide a comprehensive overview of the whole book. Both components are facilitated by the collaboration between the Junior Editor (Addition Agent $\rmA$) and the Senior Editor (Subtraction Agent $\rmS$), employing the \textit{Addition-by-Subtraction Collaboration} as depicted in \autoref{alg:add_by_sub}. \revised{For example, when constructing the glossary, the Junior Editor extracts as many terms as possible, while the Senior Editor attempts to remove generic terms.}
    
    
    \item \textbf{Tone, Style, and Target Audience}: The translation of a book is more than just a word-for-word conversion; it's a delicate process of adapting tone, style, and content to resonate with the target audience while staying true to the original text's essence. In \ours, the Senior Editor defines the tone, the style, and the target audience of the translations based on a random chapter.
    
\end{itemize}
Overall, the glossary, book summary, tone, style, and target audience together form the comprehensive translation guidelines. These guidelines are prefixed as an essential part of the prompts for all roles involved in the translation process. We present an example prompt for the Translator in \autoref{fig:guideline}. \revised{More prompts are in \autoref{sec:prompts_app}}.

\subsubsection{Execution}
\label{sec:exec}

In the execution phase, the process is divided into four sub-stages: translation, localization, proofreading, and final review. The first three sub-stages employ a collaborative strategy as detailed in \autoref{alg:trias}. Here, the Translator, Localization Specialist, and Proofreader act as Action agents $\rmP$. The Junior Editor serves as the Critique agent $\rmQ$, and the Senior Editor functions as the Judgment agent $\rmJ$. The Senior Editor also conducts the final checks before publication.

\begin{algorithm}[t]
\SetKwInOut{Input}{Input}
\SetKwInOut{Output}{Output}
\Input{
\revised{$\rmC$, the context; $\rmI$, the instruction; $\rmM$, the maximum number of iterations; $m$, the index of current iteration; $\rmP$, the Action agent; $\rmQ$, the Critique agent; $\rmJ$, the Judgment agent; $\rmF$, the feedback from the Critique agent; $\rmD$, the judgment decision;}
}
\Output{\revised{$\rmR$, the final response;}}

$\rmH \leftarrow [\rmC; \rmI]$ \quad $\triangleright$ Initialize the conversation;\\
$m \leftarrow 0$ \quad $\triangleright$ Current round;\\
\While{$m < \rmM$}{
$m \leftarrow m+1$;\\
$\rmR \leftarrow \rmP(\rmH)$ \quad $\triangleright$ Generate response;\\
$\rmF \leftarrow \rmQ(\rmH, \rmR)$ \quad $\triangleright$ Generate critiques;\\
$\rmH \leftarrow \rmH + [\rmR;\rmF]$ \quad $\triangleright$ Append $\rmR'$ and $\rmF$ to the conversation history $\rmH$;\\
\If{$m > 1$}{
$\rmD \leftarrow \rmJ(\rmC, \rmI, \rmR)$ \quad $\triangleright$ The Judgment agent $\rmJ$ evaluate the response quality;\\
\If{$\rmD = \text{TRUE}$}{\label{line:trias_exit}
Break \quad $\triangleright$ Stop iterating if the Judgment agent $\rmJ$ thinks the response is of high quality;
}
}
}
Return the final response $\rmR$;
\caption{Trilateral Collaboration}
\label{alg:trias}
\end{algorithm}
\revised{
\paragraph{Trilateral Collaboration}
We divide the collaboration in \ours into three branches, referred to as \textit{Trilateral Collaboration}:
\begin{itemize}
    \item \textbf{Action}: Executes instructions and implements required actions.
    \item \textbf{Critique}: Reviews the response and provides constructive feedback to the Action branch.
    \item \textbf{Judgment}: Makes the final decision on whether the response is satisfactory or needs further revision.
\end{itemize}
Each branch is managed by a dedicated agent as detailed in \autoref{alg:trias}. The \textit{Action agent} $\rmP$ generates a response $\rmR$ based on the context $\rmC$ and instruction $\rmI$. The \textit{Critique agent} $\rmQ$ then critiques the response $\rmR$. The \textit{Action agent} $\rmP$ can either accept the critiques and update the response or maintain the original response. Finally, the \textit{Judgment agent} $\rmJ$ evaluates the response $\rmR$ to decide if the discussion can be concluded, \textit{without requiring the conversation history}, due to the agents' limited capability of processing long-range context. Similar to \autoref{alg:add_by_sub}, we also introduce the ``early exit'' mechanism if \textit{Judgment agent} $\rmJ$ thinks the iteration process can terminate.
}
\revised{Note that the maximum number of iterations in \autoref{alg:trias} is 2.}

\paragraph{Translation, Localization, and Proofreading}

The translation stage involves three key roles: the Translator, the Junior Editor, and the Senior Editor, who collaborate to translate the book from the source language to the target language on a chapter-by-chapter basis. The process starts with the Translator (Action agent $\rmP$) translating the chapter content. The Junior Editor (Critique agent $\rmQ$) then reviews the translation, ensuring it adheres to guidelines and identifying any potential errors or areas for improvement. Lastly, the Senior Editor (Judgment agent $\rmJ$) evaluates the translation and decides if further revision is needed. Following translation, the cultural adaptation process begins. The Localization Specialist adapts the content to fit the cultural context of the target audience, ensuring it resonates well and maintains the intended meaning. Next, the Proofreader then checks for language errors. Throughout cultural adaptation and proofreading, both the Junior Editor and the Senior Editor continue to critique and evaluate the content for further refinement.

\paragraph{Final Review}

The final review is the concluding step in the editorial process. Here, the Senior Editor evaluates the translation quality and checks the flow between adjacent chapters. The review ensures each chapter is coherent and meets quality standards while maintaining smooth transitions for narrative consistency. \revised{If the translation does not meet the required standards during the final review, the steps in the execution stage are repeated to address any issues, as shown in \autoref{fig:workflow}.}

\revised{
\subsection{Discussion}
\paragraph{System Design}
Our multi-agent framework aligns with the commonly recognized translation project management process \citep{landers2001literary,perez2002translation, terhaar2019project, walker2022translation}. Besides the preparation stage, there are three core steps in the execution stage: translation, localization, and proofreading, each with its own specific aims. The translation phase leverages multi-agent collaboration to draft a coherent, context-aware translation from the source text. Although the translation from this step is already highly accurate thanks to the detailed translation guidelines, its output can still contain inaccuracies or lack cultural nuances, which is why the subsequent phases are indispensable. Next, we employ the localization phase to adapt the text more deeply to cultural, stylistic, or domain-specific norms, which is a distinct and critical step for achieving a fluent and natural translation of the content. Furthermore, the proofreading step ensures the translated text meets quality assurance standards by resolving residual errors such as typos, omissions, or inconsistencies. Together, these steps enable \ours to produce high-quality translations, like those from human translation agencies.
\paragraph{Collaboration Strategies Comparison}
In this work, we design two collaboration strategies, \textit{Addition-by-Subtraction Collaboration} and \textit{Trilateral Collaboration}. These two strategies are designed based on the tasks the agents are dealing with and the capability of LLM backbone. They are not directly comparable. When constructing the translation glossary, we observe that the \textit{Addition agent} is very likely to include some generic words into the list, so we introduce the \textit{Subtraction agent} to remove these generic words. The iteration stops until the translation glossary does not update. However, this strategy is not suitable for translation, localization, and proofreading, as adding and removing content can significant degrade the translation quality. 
Conversely, \textit{Trilateral Collaboration} cannot replace \textit{Addition-by-Subtraction Collaboration} either. In our preliminary study, we observe that the \textit{Critique agent} and the \textit{Judgment agent} often fail to exclude those generic terms if we leverage \textit{Trilateral Collaboration} to construct the translation glossary, which make translation guideline overly lengthy. Long context modeling is one of the key challenges of current LLMs. An overly lengthy translation guideline can degrade the translation quality. 
}
\section{Experimental Setup}
In this work, our experimental setup primarily follows the WMT2023 shared task on discourse-level literary translation (DLLT) \citep{wang-etal-2023-findings,wang-etal-2024-findings}. The following sections introduce the baselines and datasets (\autoref{sec:baselines}), and evaluation approaches (\autoref{sec:evaluation}) used in our study.

\subsection{Baselines and Datasets}
\label{sec:baselines}
We leverage the state-of-the-art LLM \gptfour as the backbone of our agents,\footnote{\texttt{gpt-4-1106-preview}} and compare our approach with the unconstrained systems in WMT2023 shared task on DLLT, including \model{Llama-MT} \citep{duextrapolation}, \model{gpt-4} \citep{DBLP:journals/corr/abs-2303-08774}, \model{Google Translate}, \model{DUT} \citep{zhao-etal-2023-dutnlp}, and \model{HW-TSC} \citep{xie-etal-2023-hw}. \revised{We also involve \gptfour, \gptomini,\footnote{\texttt{gpt-4o-mini-2024-07-18}} and \gpto,\footnote{\texttt{gpt-4o-2024-08-06}} as our baselines.}

In this work, we only leverage the official test set of WMT2023 shared task on DLLT for evaluation. The official test set is collected from 20 web novels, each of which consists 20 consecutive chapters, totaling 240 chapters. Each chapter contains approximately 1,404 English words on average. The test set contains two references: \model{Reference 1} is translated by human translators and \model{Reference 2} is built by manually aligning bilingual text in web page.

\subsection{Evaluation}
\label{sec:evaluation}
In this work, we employ two evaluation approaches: Standard Evaluation (\autoref{sec:standard_eval}) and Preference Evaluation (\autoref{sec:prefer_eval}).

\subsubsection{Standard Evaluation}
\label{sec:standard_eval}
Following \citet{wang-etal-2023-findings}, we use \dbleu \citep{papineni-etal-2002-bleu, post-2018-call, liu-etal-2020-multilingual-denoising} to evaluate the translation quality,\footnote{\texttt{nrefs:2|case:mixed|eff:no|tok:13a|\\smooth:exp|version:2.3.1}} as the translations may not strictly align with the source text on a sentence-by-sentence basis. To compute the \dbleu score, we concatenate all the chapter translations into a single document for evaluation.
\revised{Both references in the test set are used for computing \dbleu.}
\revised{Furthermore, we use \gemba \citep{kocmi-federmann-2023-large} with \gpto as the evaluator to assess translation quality chapter by chapter. It is important to note that recent neural MT metrics, such as \model{COMET} \citep{rei-etal-2020-comet}, are typically designed for sentence-level evaluation and have restricted context sizes. For instance, \model{COMET} has a context window of only 512 tokens, while \model{DocCOMET} \citep{vernikos-etal-2022-embarrassingly} supports slightly longer contexts but requires sentence-by-sentence alignment.}
\revised{Additionally, we also conduct the bootstrap resampling and significance testing \citep{koehn-2004-statistical} to better understand the results.}


\subsubsection{Preference Evaluation}
\label{sec:prefer_eval}

Acknowledging the multifaceted nature of literary texts is essential, as they do not have a single, universal translation. Standard translation evaluations, which rely on comparisons to a standard reference, fall short in capturing this complexity. \revised{Following \citet{thai-etal-2022-exploring}, we segment each chapter into approximately 150-word sections based on the story development and then utilize both human evaluators and large language models (LLMs) to assess translations based on their preferences. We detail our evaluation methods in this section.}

\paragraph{Monolingual Human Preference (MHP)}
When reading a translated book, understanding the original language is unnecessary for the target audience. Thus, readers should prefer a better translation without referencing the original text. Human evaluators compare pairs of translation segments describing the same part of the story and select their preferred translation using the provided interface in \autoref{sec:mhp_app}. To ensure context is considered, evaluators review all segments within a chapter in their original order, as segments may rely on preceding information. 

\revised{
\paragraph{Implementation Details of MHP}
In this work, we collect human preferences on translation segments through SurveyMonkey.\footnote{\url{https://www.surveymonkey.com/}} We only recruit evaluators from the United States to minimize potential impacts of demographics. Each translation pair is evaluated by at least 10 people and costs us \$0.30 USD per annotation. During the annotation process, we randomly swap the positions of translation segments for each comparison to avoid the positional bias from human annotators. We filter out possible low-quality responses or human evaluators based on following criteria:
\begin{itemize}
    \item Being labeled as low quality by SurveyMonkey's response quality model;
    \item Selecting the same option for all selections;
    \item Taking less than 20 seconds per annotation.
\end{itemize}
After filtering, we collect at least 5 responses per pair. Furthermore, we aggregate the human evaluations using majority voting, where the most selected option is considered the final result. If two translations receive the same number of votes, we record the result as ``No Preference'' (Tie). 
}


\begin{figure}[t]
    \centering
    \footnotesize
    \begin{Verbatim}[frame=single, fontsize=\tiny, breaklines=true, breakanywhere=true]
[The start of source]
[$src_lang]: $src
[The end of source]

[The start of assistant 1's translation]
[$tgt_lang]: $asst1
[The end of assistant 1's translation]

[The start of assistant 2's translation]
[$tgt_lang]: $asst2
[The end of assistant 2's translation]

We would like to request your feedback [TRUNCATED]
    \end{Verbatim}
    \caption{The prompt used for bilingual LLM preference evaluation.}
    \label{fig:gpt4eval_prompt}
\end{figure}

\revised{
\paragraph{Bilingual LLM Preference (BLP)}
Recent works demonstrate that the reference translations are likely to be of low quality \citep{freitag-etal-2023-results, DBLP:journals/corr/abs-2401-08417}. \citet{kocmi-federmann-2023-gemba} demonstrate that \model{gpt-4} is capable of accurately estimating translation quality without the need for human reference translations. Hence, we require \model{gpt-4-0125-preview} to determine which translation segment is better, without providing the reference translations, as shown in \autoref{fig:gpt4eval_prompt}. Each segment pair is assessed in both forward and reversed directions.
}



\paragraph{Evaluation Metrics}
For both MHP and BLP, we use the winning rate (\%), which is the percentage of instances where a model's generated chapter is preferred by either the human evaluators or the LLM, to measure the model performance.

\revised{
\paragraph{Comparison with Classical Human Evaluation}
Classical human evaluation methods, such as the Multidimensional Quality Metrics (MQM) framework \citep{burchardt-2013-multidimensional}, are typically designed for sentence-level MT. However, \ours translates the source text chapter by chapter. It is highly challenging for the annotators to perform MQM evaluations on ultra-long texts like our test examples, making the human evaluation process extraordinarily slow and expensive. In contrast, the Monolingual Human Preference (MHP) evaluation method offers several advantages. First, MHP does not require bilingual professional annotators. Instead, it relies on monolingual annotators from the target audience, simplifying the annotator recruitment process and enabling larger-scale human evaluations. Second, unlike classical human evaluation methods, MHP does not provide the source text to annotators. Instead, annotators are asked to select their preferred translations, allowing them to focus on text attributes beyond translation errors, such as naturalness, fluency, and other stylistic qualities. Furthermore, recent research has demonstrated the effectiveness of preference-based evaluation in estimating translation quality \citep{thai-etal-2022-exploring, he-etal-2024-exploring}. These distinctions make MHP a more suitable evaluation method for our use case.
}


\begin{table}[t]
\centering
\small
\begin{tabular}{lcc}
\toprule
                   & \dbleu $\uparrow$ & \gemba $\uparrow$ \\ \midrule
\model{Reference 1}        & ---    & 85.4\rlap{$_{\pm 0.5}$}   \\
\model{Reference 2}        & ---    & 82.6\rlap{$_{\pm 4.5}$}   \\ \midrule
\model{Llama-MT}   & 43.1   & ---    \\
\model{gpt-4-0613} & 43.7   & ---    \\
\model{Google}     & 47.3   & ---    \\
\model{DUT}        & 50.2   & ---    \\
\model{HW-TSC}     & 52.2   & ---    \\ \midrule
\gptfour      & 47.8\rlap{$_{\pm 2.6}$}   & 85.8\rlap{$_{\pm 0.5}$}   \\
\gptomini & 46.1\rlap{$_{\pm 2.5}$} & 86.9\rlap{$_{\pm 0.4}\dagger$} \\
\gpto & 46.3\rlap{$_{\pm 2.6}$} & 87.1\rlap{$_{\pm 0.3}\dagger$} \\ \midrule
\ours              & 25.0\rlap{$_{\pm 2.4}$}   & 87.7\rlap{$_{\pm 0.2}\dagger$}   \\ \bottomrule
\end{tabular}
\caption{
\dbleu and \gemba scores with standard deviation on WMT2023 DLLT test set. $\uparrow$ indicates higher is better. \gemba score is in the range from 0 to 100. $\dagger$ indicates the improvement is statistically significant against \model{Reference 1} at the significance level $\alpha=0.05$ according to \citet{koehn-2004-statistical}.
}
\label{tab:main_auto}
\end{table}

\section{Main Results}
\label{sec:main_results}

We present the main results on standard evaluation and preference evaluation in this section.
\revised{
\paragraph{Standard Evaluation Results}
We present the automatic evaluation results in \autoref{tab:main_auto}. Our approach performs poorly in terms of the \dbleu metric, achieving the lowest scores among the compared methods. However, it is important to note that \dbleu has limitations as an evaluation metric, as it primarily focuses on surface-level similarity and may not fully capture the quality and coherence of the generated text. Although the \dbleu score of \ours is significantly lower than that of \model{gpt-4}, \ours achieves a higher \gemba score. The \gemba results suggest that \ours delivers better overall quality despite its lower performance on \dbleu.
}


\begin{figure}

\centering
\begin{minipage}{.45\textwidth}

\centering
\scriptsize
\begin{tikzpicture}
\begin{axis}[
  xbar stacked,
  height=3cm,
  width=0.8\textwidth,
  enlarge x limits={abs=0.2cm},
  enlarge y limits={abs=0.4cm},
  bar width=7pt,
  xmin=0, xmax=100,
  legend style={
    at={(0.5,-0.7)}, 
    anchor=north,
    legend columns=-1,
    font=\tiny
    },
  xlabel={Percentage (\%)},
  symbolic y coords={
    Reference 1,
    GPT-4,
    },
  ytick=data,
  nodes near coords,
  every node near coord/.append style={font=\tiny},
  tick label style={font=\scriptsize},
  ]
\addplot [xbar, fill=green!30, draw=green] coordinates {
(52.1,Reference 1)
(55.6,GPT-4)
};
\addplot [xbar, fill=blue!30, draw=blue] coordinates {
(8.4,Reference 1)
(6.8,GPT-4)
};
\addplot [xbar, fill=red!30, draw=red] coordinates {
(39.5,Reference 1)
(37.6,GPT-4)
};
\legend{\texttt{\ours wins}, \texttt{Tie}, \texttt{\ours loses}}
\end{axis}
\end{tikzpicture}
\caption{
    Monolingual Human Preference evaluation results. GPT-4 indicates \model{gpt-4-1106-preview}.
}
\label{fig:mhp_eval}

\end{minipage}%
\hspace{8pt}
\begin{minipage}{.45\textwidth}

\centering
\scriptsize
\begin{tikzpicture}
\begin{axis}[
  xbar stacked,
  height=3cm,
  width=0.8\textwidth,
  enlarge x limits={abs=0.2cm},
  enlarge y limits={abs=0.4cm},
  bar width=7pt,
  xmin=0, xmax=100,
  legend style={
    at={(0.5,-0.7)}, 
    anchor=north,
    legend columns=-1,
    font=\tiny
    },
  xlabel={Percentage (\%)},
  symbolic y coords={
    Reference 1,
    GPT-4,
    },
  ytick=data,
  nodes near coords,
  every node near coord/.append style={font=\tiny},
  tick label style={font=\scriptsize},
  ]
\addplot [xbar, fill=green!30, draw=green] coordinates {
(66.2,Reference 1)
(55.9,GPT-4)
};
\addplot [xbar, fill=blue!30, draw=blue] coordinates {
(2.9,Reference 1)
(6.7,GPT-4)
};
\addplot [xbar, fill=red!30, draw=red] coordinates {
(30.9,Reference 1)
(37.4,GPT-4)
};
\legend{\texttt{\ours wins}, \texttt{Tie}, \texttt{\ours loses}}
\end{axis}
\end{tikzpicture}
\caption{
    Bilingual LLM Preference evaluation results. GPT-4 indicates \model{gpt-4-1106-preview}.
}
\label{fig:blp_eval}

\end{minipage}
\end{figure}

\paragraph{Preference Evaluation Results}

We compare the performance of our \ours model with \model{Reference 1} and \model{gpt-4-1106-preview} using monolingual human preference (MHP) and bilingual LLM preference (BLP) evaluations. The results, presented as winning rates, are shown in \autoref{fig:mhp_eval}. The translations produced by \ours are marginally preferred by human evaluators compared to both \model{Reference 1} and \model{gpt-4-1106-preview}. \revised{The Cohen's $\kappa$ for the MHP between \model{Reference 1} and \ours is 0.17, while the Cohen's $\kappa$ for the MHP between \model{gpt-4-1106-preview} and \ours is 0.11.} Additionally, the translations generated by \ours are also more favored by \model{gpt-4-0125-preview} compared to the other models, as shown in \autoref{fig:blp_eval}. Both evaluation methods demonstrate that \ours can generate superior translations compared with the human translators and \model{gpt-4-1106-preview}.



\section{Analysis}
\label{sec:analysis}


\begin{table}[t]
\centering
\small
\setlength{\tabcolsep}{8pt}
\begin{tabular}{lcc}
\toprule
                           & d-BLEU                  & \gemba                  \\ \midrule
\gptfour & 47.8\rlap{$_{\pm 2.6}$} & 85.8\rlap{$_{\pm 0.5}$} \\ \midrule
\ours                      &                         &                         \\
\quad -- Translation       & 28.8\rlap{$_{\pm 2.3}$} & 86.4\rlap{$_{\pm 0.2}$} \\
\quad -- Localization      & 25.5\rlap{$_{\pm 2.5}$} & 87.3\rlap{$_{\pm 0.3}$} \\
\quad -- Proofreading      & 25.0\rlap{$_{\pm 2.4}$} & 87.7\rlap{$_{\pm 0.2}$} \\ \bottomrule
\end{tabular}
\caption{
d-BLEU and \gemba scores with standard deviation given by each stage in \ours on WMT2023 DLLT test set.
Note that the ``proofreading'' translation is the final translation of \ours.}
\label{tab:step}\label{tab:why_fail}
\end{table}
\paragraph{What Causes \ours to ``Fail'' in Terms of \dbleu?}
As shown in \autoref{tab:main_auto}, the translation produced by \ours achieves the lowest \dbleu score among the compared methods. To investigate the reasons behind this, we evaluate the output of each stage in the \ours workflow using the official references from the WMT2023 DLLT test set. 
\revised{
As shown in \autoref{tab:why_fail}, although \ours achieves lower \dbleu scores, the translation quality improves further in terms of \gemba after each step. This demonstrates the effectiveness of each step in \ours and highlights the limitations of \model{BLEU}. Furthermore, the translation guideline is likely to be the primary contributor to the final translation quality, as the largest decline in \dbleu comes from the Translation step of \ours, compared to \gptfour. Additionally, we conduct an experiment that rephrases the references of FLORES \citep{DBLP:journals/corr/abs-2207-04672} and present the results in \autoref{sec:rephrase}.
}


\begin{table*}[t]
\centering
\small
\setlength{\tabcolsep}{8pt}
\begin{tabular}{lccccccccc}
\toprule
                           & Overall & VG   & EF   & SR   & CR   & F    & SF   & HT   & FR   \\ \midrule
\multicolumn{10}{l}{\textit{Monolingual Human Preference}}                                            \\
\gptfour & 55.6    & 64.5 & 68.2 & 63.3 & 44.6 & 68.2 & \underline{39.1} & 48.0 & \textbf{77.8} \\
\model{Reference 1}        & 52.1    & \textbf{67.7} & 63.6 & 56.7 & 42.9 & 63.6 & \underline{37.0} & 40.0 & 66.7 \\ \midrule
\multicolumn{10}{l}{\textit{Bilingual LLM Preference}}                                                \\
\gptfour & 55.9    & \textbf{74.1} & 56.8 & 58.3 & 47.3 & 70.5 & 47.8 & \underline{34.0} & 66.7 \\
\model{Reference 1}        & 66.2    & \textbf{88.7} & 59.1 & 70.0 & 54.5 & 83.0 & \underline{53.3} & 62.0 & 61.1 \\ \bottomrule
\end{tabular}
\caption{
The breakdown winning rate of \ours against \gptfour and \model{Reference 1}.
Best results in each row are highlighted in \textbf{bold}.
Worst results in each row are highlighted in \underline{underline}.
}
\label{tab:breakdown_win}
\end{table*}
\paragraph{Strengths and Weaknesses of \ours regarding Genres}
The test examples span a variety of genres, including Video Games (VG), Eastern Fantasy (EF), Sci-fi Romance (SR), Contemporary Romance (CR), Fantasy (F), Science Fiction (SF), Horror \& Thriller (HT), and Fantasy Romance (FR). We present a detailed analysis of the performance of our model \ours, across these categories in \autoref{tab:breakdown_win}. Our observations indicate that \ours excels in domains that demand extensive domain-specific knowledge, such as historical contexts and cultural nuances. These areas often pose significant challenges for translators. Meanwhile, \ours underperforms in contemporary domains, which do not require as much specialized knowledge. This trend underscores \ours's strengths and weaknesses. 

\begin{table}[t]
\centering
\small
\begin{tabular}{lcc}
\toprule
                           & \model{MATTR} $\uparrow$ & \model{MTLD} $\uparrow$ \\ \midrule
\model{Reference 1}        & 80.9                  & \phantom{0}89.1        \\
\model{gpt-4-1106-preview} & 81.5                  & \phantom{0}94.9        \\ \midrule
\ours                      &                       &                        \\
\quad -- Translation       & 83.5                  & 117.0                  \\
\quad -- Localization      & 83.6                  & 119.4                  \\
\quad -- Proofreading      & 83.6                  & 119.4                  \\ \bottomrule
\end{tabular}
\caption{
Linguistic diversity in terms of \model{MATTR} (up-scaled by $\times 100$) and \model{MTLD}.
$\uparrow$ indicates higher is better.
}
\label{tab:lingdiv}
\end{table}

\paragraph{Linguistic Diversity}
Linguistic diversity in literary texts is critical for enriching the reading experience. To quantify the linguistic diversity of the translation, we leverage two metrics: the Moving-Average Type-Token Ratio (MATTR) \citep{covington2010cutting} and the Measure of Textual Lexical Diversity (MTLD) \citep{mccarthy2010mtld}. As shown in \autoref{tab:lingdiv}, assisted by our translation guidelines, our initial translation significantly improves linguistic diversity compared to the reference text. Moreover, the localization step further enhances linguistic diversity, while the proofreading step does not affect it. These results demonstrate the effectiveness of our approach in preserving and enhancing the richness of language in the translated literary work.

\begin{table}[t]
\centering
\small
\setlength{\tabcolsep}{3pt}
\begin{tabular}{lccc}
\toprule
                           & \model{MTLD} $\uparrow$  & \model{BLP\textsubscript{Ref}} $\uparrow$ & \model{BLP\textsubscript{GPT}} $\uparrow$ \\ \midrule
\ours                      &       &                        &                        \\
\quad + None               & 113.8 & 48.2                   & 40.1                   \\
\quad + Minimum            & 117.9 & 59.6                   & 48.2                   \\
\quad + LangSpec & 118.8 & 64.4                   & 52.7                   \\
\quad + Vivid              & 119.4 & 66.2                   & 55.9                  \\ \bottomrule
\end{tabular}
\caption{
Analysis on agent profile design.
MTLD measures the linguistic diversity.
\model{BLP\textsubscript{Ref}} and \model{BLP\textsubscript{GPT}} indicate the winning rate (\%) of Bilingual LLM preference against \model{Reference 1} and \gptfour. $\uparrow$ indicates higher is better.
}
\label{tab:profile_analysis}
\end{table}
\paragraph{Agent Profile Design}
Recent work demonstrates that diverse role-playing profiles can effectively enhance model performance \citep{chan2024scaling}. We design agent profiles with detailed and rich personas, as described in \autoref{sec:overview}. To validate the effectiveness of our design choices on agent profiles, we categorize the agent profiles into four groups based on their level of detail:
\begin{itemize}
    \item \textbf{None}: The role-playing profile is not used.
    \item \textbf{Minimum}: The profile includes only the job title of the agent, for example, \texttt{You are a Senior Editor}.
    \item \textbf{Language-Specified (LangSpec)}: The profile includes the job title and language skills of the agent, for example, \texttt{You are a Senior Editor, specializing in English and Chinese}.
    \item \textbf{Vivid}: The profile includes detailed and rich personas as described in \autoref{sec:overview}.
\end{itemize}
We present the results in \autoref{tab:profile_analysis} and demonstrate that incorporating a role-playing message can significantly enhance model performance. We observe that agent profile with more details typically lead to better performance. Our \textbf{Vivid} agent profile, which includes the most detailed and colorful information, achieves the best performance.

\begin{figure}[t]

\centering
    \begin{tikzpicture}
    \begin{axis}[
        width=0.45\textwidth,
        height=5cm,
        xlabel={$\rmM$},
        ylabel={\gemba},
        xtick={1,2,3,4,5},
        ymin=84.5,ymax=88.5,
        legend style={
            at={(00.49,-0.3)}, 
            anchor=north,
            legend columns=2,
            cells={anchor=west},
            font=\small
        },
        grid=major,
    ]
    
    \addplot[mark=*, blue] coordinates {
        (1,86.1)
        (2,87.7)
        (3,87.5)
        (4,87.6)
        (5,87.5)
    };
    \addlegendentry{$\rmM$ in \autoref{alg:add_by_sub}}

    \addplot[mark=square, red] coordinates {
        (1,86.2)
        (2,87.7)
        (3,87.2)
        (4,87.4)
        (5,87.4)
    };
    \addlegendentry{$\rmM$ in \autoref{alg:trias}}

    \draw [dashed,green] (axis cs:0,85.4) -- (axis cs:10,85.4);
    \node[anchor=west] at (0.5,85.3) {\scriptsize \model{Reference 1}};

    \draw [dashed,red] (axis cs:0,85.8) -- (axis cs:10,85.8);
    \node[anchor=west] at (1.2,86.0) {\scriptsize \gptfour};



    \end{axis}
    \end{tikzpicture}

\caption{
\gemba scores given by \ours with regard to $\rmM$ in \autoref{alg:add_by_sub} and \autoref{alg:trias}. The green and red dashed lines indicate the \gemba scores of \model{Reference 1} and \gptfour.
}
\label{fig:round_gemba}
\end{figure}

\revised{
\paragraph{Impact of Iterations}
we conduct additional experiments to analyze the effect of varying $\rmM$ and present the results in \autoref{fig:round_gemba} for \autoref{alg:add_by_sub} and \autoref{alg:trias}, respectively. Our findings indicate that for $\rmM > 2$, the improvements in translation quality become minimal. This behavior can be attributed to our ``early exit'' design, where the iteration terminates early if the glossary list and summary converge quickly (at \autoref{line:add_by_sub_exit} of \autoref{alg:add_by_sub}) or if the Judgment agent $\rmJ$ in \autoref{alg:trias} determines that the output is of high quality (at \autoref{line:trias_exit} of \autoref{alg:trias}). For ultra-long literary texts, the limitations of current LLMs in processing long contexts become more pronounced, and increasing $\rmM$ beyond 2 can sometimes degrade translation quality. Furthermore, thanks to our ``early exit'' design, the runtime of \ours increases by only about 10\% when $\rmM > 2$.
}

\begin{table}[t]
\centering
\small
\setlength{\tabcolsep}{3.5pt}
\begin{tabular}{lcccc}
\toprule
                               & \multicolumn{2}{c}{En-Zh}  & \multicolumn{2}{c}{En-De}  \\ \cmidrule(rl){2-3} \cmidrule(l){4-5}
                               & \dbleu & \model{C.K.} & \dbleu & \model{C.K.} \\ \midrule
\gptfour     & 42.7   & 79.3              & 25.5   & 70.5              \\
\gptomini & 42.9   & 79.3              & 26.6   & 71.2              \\
\gpto      & 45.6   & 80.5              & 28.3   & 72.4              \\ \midrule
\ours                          & 29.3   & 75.5              & 24.0   & 72.4              \\ \bottomrule
\end{tabular}
\caption{
\dbleu and \model{CometKiwi} scores on English-Chinese and English-German translation tasks of WMT2024 General MT test set. \model{C.K.} indicates \model{CometKiwi}.
}
\label{tab:wmt}
\end{table}
\revised{
\paragraph{Short-Text Translation}
We conduct additional experiments on English-Chinese and English-German translations using the WMT2024 General MT tasks. Due to budget constraints, we randomly selected 200 test examples from the entire test set, and the results are presented in \autoref{tab:wmt}. When translating shorter texts, \ours performs worse than \gptomini and \gpto, as expected. Upon manual inspection of the translations and the conversations between agents, we found that the agents often fail to correctly determine the style, tone, and target audience when provided with only a short text. Consequently, the translations produced by \ours frequently misalign with the style and tone of the source and reference sentences and suffer from overtranslation issues. We present a case study for the short text translation in \autoref{sec:case_study}.
}

\begin{table}[t]
\centering
\small
\setlength{\tabcolsep}{6pt}
\begin{tabular}{lccc}
\toprule
                         & \model{G.D.} & Runtime  & Cost      \\ \midrule
\model{Reference 1}                  & 85.4   & a few weeks           & est. 40K             \\ \midrule
\gptfour   & 85.8   & \phantom{0}1.1 & \phantom{0}42.1 \\
\ours                       & 87.7   & 13.1           & 487.3           \\ \bottomrule
\end{tabular}
\caption{
Comparison among \model{Reference 1}, \gptfour, and \ours. \model{G.D.} indicates \gemba. The runtime (in hours) and cost (in USD) are calculated based on the entire test set.
}
\label{tab:cost}
\end{table}
\revised{
\paragraph{Cost Analysis}
We provide a more detailed evaluation of \model{Reference 1},\footnote{The minimum rate for translation services is \$0.12 USD per word. Source: \url{https://tinyurl.com/bdze92xr}} \gptfour, and \ours in terms of translation quality, runtime, and financial costs, as shown in \autoref{tab:cost}. Our analysis reveals that \ours introduces significant overhead in both runtime and costs compared to \model{gpt-4-1106-preview}. However, it remains substantially more cost-effective and efficient than professional human translators.
}

\section{Case Study}
\label{sec:case_study}
\revised{
In this section, we explore case studies on cultural adaptation and shorter texts. We also discuss a case on undertranslation and include comments from professional translators in \autoref{sec:more_cases}. 
}

\begin{CJK*}{UTF8}{gbsn}

\begin{table}[t]
\centering
\tiny
\setlength{\tabcolsep}{3pt}
\begin{tabular}{@{}lp{0.37\textwidth}@{}}
\toprule
Source Text              & 罗德抬起头来，正好看见一个中年男子推门走进来，他穿着冒险者的皮甲，一头鲜红的长发随意的向后梳理着， 看见罗德，男子微微一笑。 “你好，先生，我是星月佣兵团的团长，卡特。”                                                    \\ \midrule
\model{Reference 1}        & Rhode looked up and saw a middle-aged man pushing the door open. He wore an adventurer's leather armor and casually combed back his long red hair. The man smiled when he saw Rhode. \textcolor{red}{"Hello, sir. I am the Guild Leader of the Star Moon Mercenaries, Carter."} \\ \midrule
\gptfour & Rhode looked up and saw a middle-aged man entering through the door. He was dressed in an adventurer's leather armor, with a head of bright red hair casually combed back. Seeing Rhode, the man smiled slightly. \textcolor{red}{"Hello, sir, I am the leader of the Star Moon Mercenary Group, Carter."}   \\ \midrule
\multicolumn{2}{c}{\cellcolor{gray!30}\ours}  \\
\model{Translation} & Rhode looked up, only to behold a middle-aged man making his entrance. Clad in the leather armor of an adventurer, his fiery red hair swept back carelessly. Upon noticing Rhode, the man's lips curled into a modest smile. \textcolor{red}{"Greetings, sir. I am the leader of the Star Moon Group, Carter."}   \\ \midrule
\model{Localization} & \textcolor{red}{Rod} looked up to see a middle-aged man entering. The man was dressed in the leather armor typical of adventurers, his fiery red hair casually swept back. Spotting \textcolor{red}{Rod}, the man offered a modest smile.. \textcolor{blue}{``Hello, sir. I am Carter, the leader of} \textcolor{red}{the Star Moon Group}.''   \\ \midrule
\model{Proofreading}                      & \textcolor{blue}{Rhode} looked up to see a middle-aged man entering. The man was dressed in the leather armor typical of adventurers, his fiery red hair casually swept back. Spotting \textcolor{blue}{Rhode}, the man offered a modest smile. \textcolor{blue}{``Hello, sir. I am Carter, the leader of the Star Moon Mercenary Corps.''}  \\ \bottomrule
\end{tabular}
\caption{
Case study on cultural adaptation.
\textcolor{red}{The text in red} indicates that the translation is accurate in meaning but not in cultural context.
\textcolor{blue}{The text in blue} indicate that the translation is accurate both in meaning and in cultural context.
}
\label{tab:case_culture}
\end{table}
\end{CJK*}
\paragraph{Cultural Adaptation}
In Chinese, job titles are typically placed before a person's name, whereas in English, job titles usually come after the person's name. This order reflects differing linguistic and cultural conventions regarding the structuring of personal information in the two languages. 
\revised{
As demonstrated in \autoref{tab:case_culture}, both \model{Reference 1} and \gptfour fail to correctly adjust the order of names and job titles, thus not adhering to the cultural norms expected in the target language. In contrast, \ours accurately reflects this cultural context in its translation after the translation-localization-proofreading workflow. In the initial translation phase, the output of \ours is of high quality but contains errors related to cultural adaptation and terminology usage. During the localization step, the cultural adaptation issues are resolved, but errors in terminology persist. Finally, the proofreading step addresses the remaining terminology issues. This example effectively demonstrates the utility of each step in the \ours process. This emphasizes the capability of \ours to provide translations that are culturally appropriate, ensuring an immersive reading experience for readers.
}



\begin{CJK*}{UTF8}{gbsn}

\begin{table}[t]
\centering
\tiny
\setlength{\tabcolsep}{3pt}
\begin{tabular}{@{}lp{0.37\textwidth}@{}}
\toprule
Source Text              & Adapt the old, accommodate the new to solve issue                                                    \\ \midrule
\model{Reference}        & 适应旧的，容纳新的，积极解决问题。  (Adapt to the old, accommodate the new, and actively solve problems) \\ \midrule
\gptfour & 适应旧的，接纳新的以解决问题。(Adapt to the old and embrace the new to solve problems.)   \\ \midrule
\gptomini & 适应旧的，容纳新的，以解决问题。(Adapt the old and accommodate the new to solve problems.)   \\ \midrule
\gpto & 适应旧事物，接纳新事物以解决问题。(Adapt to old things and accept new things to solve problems.)   \\ \midrule
\ours                      & 传承与创新并重，以应对当代挑战。 (Inheritance and innovation are equally important to meet contemporary challenges)  \\ \bottomrule
\end{tabular}
\caption{
Case study on short-text translation. The texts within parentheses are the back-translations provided by \model{Google Translate}.
}
\label{tab:wmt_case}
\end{table}
\end{CJK*}
\revised{
\paragraph{Short-Text Translation}
As mentioned in \autoref{sec:analysis}, the translations produced by \ours frequently misalign with the style and tone of the source and reference sentences due to the short context. For instance, as shown in \autoref{tab:wmt_case}, the translations generated by \gptfour, \gptomini and \gpto are almost identical to the reference translation. In contrast, the translation produced by \ours are more polished and idiomatic in Chinese but diverge somewhat from the literal meaning of the original text. 
}

\section{Conclusion}

\revised{
We introduce \ours, a novel multi-agent framework for literary translation inspired by the structured workflows of human translation companies. The framework divides the process into preparation and execution stages, assigning distinct roles to agents. To evaluate its effectiveness, we propose innovative methods, including Monolingual Human Preference (MHP) and Bilingual LLM Preference (BLP). Our findings show that the framework excels in translating long texts and capturing the nuances of literary works, though challenges with shorter texts highlight areas for improvement. Insights from agent profiling and collaboration strategies offer guidance for refining future translation systems.  
In summary, our work paves the way for new directions in developing translation systems that combine LLMs with the best practices of human translation.
}

\section*{Acknowledgments}

We sincerely thank the action editor and reviewers for their valuable time, insightful feedback, and constructive suggestions. Their thoughtful comments have greatly contributed to improving the quality and clarity of our work. We deeply appreciate their efforts in helping us refine this manuscript.

\bibliography{iclr2024_conference}
\bibliographystyle{acl_natbib}

\clearpage
\appendix



\section{Interface for Monolingual Human Preference}
\label{sec:mhp_app}

We present the user interface for MHP in \autoref{fig:survey_interface}.


\revised{
\section{Rephrasing Experiment}
\label{sec:rephrase}
We acknowledge that a drop from 47.8 to 25.0 in d-BLEU may intuitively suggest poorer translation performance. However, we argue that this significant decline in BLEU does not necessarily indicate poorer translation quality but rather reflects the inherent limitations of BLEU as a metric. BLEU primarily measures surface-level n-gram overlap, which often fails to capture deeper semantic and stylistic qualities, especially in nuanced domains like literary translation. To illustrate this limitation, we conduct an additional experiment on the FLORES Chinese-English test set. First, we used \gpto to generate translations and then rephrase these translations to use entirely different vocabulary while preserving the intended meaning. We evaluate the original references, the translations generated by \gpto, and their rephrased versions using BLEU and \model{CometKiwi}.\footnote{BLEU signature: \texttt{nrefs:1|case:mixed|eff:no\\|tok:13a|smooth:exp|version:2.3.1} and \model{CometKiwi} signature: \texttt{Unbabel/wmt22-cometkiwi-da}} As shown in \autoref{tab:rephrase}, the BLEU scores decline significantly after rephrasing, while the \model{CometKiwi} scores remaine almost unchanged. These results highlight BLEU's inability to account for semantic preservation when lexical variation occurs. Therefore, \textbf{the large drop in BLEU does not suggest poor translation performance by our approach but rather underscores the BLEU's limitations in evaluating nuanced translations and poor diversity of references, as discussed by \citet{freitag-etal-2020-bleu}.}
}
\begin{figure}[t]

    \begin{center}
    \begin{Verbatim}[frame=single, fontsize=\tiny, breaklines=true, breakanywhere=true]
Q: Which of the following writing style do you prefer?

[x] Chapter 455: Turnaround 3 "Allow me to demonstrate the sensing of Formless Fluctuation; it's remarkably straightforward," interjected another sorcerer, a smile evident in his voice. "Your assistance is appreciated," Lin Sheng responded, offering a nod of gratitude. Time was of the essence in finding the remaining Fragments. He had initially planned to conquer an array of Great Evil Spirits to amass substantial reserves of pure soul power. Yet, the present opportunity necessitated an immediate and decisive acquisition. Promptly, the sorcerer leader brought Lin Sheng to a daunting Evil Spirit Gate. Both extended their hands, gently touching the gate's enigmatic frame, eyes closed as one. The leader rapidly employed his Special Ability to establish a Spatial Foundation, thus setting a Coordinate Code.

[ ] Chapter 455 Reversion 3 "This is to let you feel the fluctuation of aura. It's really simple." Another Warlock couldn't help but interrupt with a smile. "Then I'll have to trouble you." Lin Sheng nodded. He needed to find the other fragments as soon as possible. Originally, he had planned to conquer more evil spirits and obtain more pure soul power. But now that he encountered such an opportunity, the most important thing for him was to get it as soon as possible. Soon, the Warlock Commander led Lin Sheng to an Evil Spirit Gate. The two reached out, touched the frame of the Evil Spirit Gate at the same time, and closed their eyes. The Warlock Commander quickly used his ability to build the space base as a coordinate.

[ ] No Preference
    \end{Verbatim}
    \end{center}
    \caption{
    The user interface for Monolingual Human Preference (MHP).
    \texttt{[x]} indicates the selection of human evaluator.
    }
    \label{fig:survey_interface}

\end{figure}
\begin{table}[t]
\centering
\small
\begin{tabular}{lcc}
\toprule
                    & BLEU & \model{CometKiwi} \\ \midrule
Reference 1          & 100.0  & 82.7              \\
Rephrased Reference 1 & \phantom{0}12.5 & 82.7              \\
\gpto               & \phantom{0}29.4 & 84.9              \\
Rephrased \gpto     & \phantom{0}11.5 & 82.8              \\ \bottomrule
\end{tabular}
\caption{FLORES Chinese-English translation results given by the reference, rephrased reference, \gpto, and rephrased \gpto. The \model{CometKiwi} is scaled up by $\times$ 100.}
\label{tab:rephrase}
\end{table}

\section{More Cases}
\label{sec:more_cases}
In this section, we present additional case study on undertranslation and comments from professional translators.

\begin{CJK*}{UTF8}{gbsn}

\begin{table*}[t]
\centering
\tiny
\begin{tabular}{@{}lp{0.8\textwidth}@{}}
\toprule
Original Text              & 她招来女仆带叶琛和程安雅下去洗漱，小奶包虽然很想跟着去， 不过他还是留在这里，白夜作势就要揍人了，小奶包赶紧拉着他的袖子。 “白夜，你能有办法救我爹地妈咪吗？”小孩子的眼睛很亮，如两颗黑葡萄镶嵌在白嫩的脸上，充满了期盼，仿佛白夜一摇头，他眸中的亮光就会黯淡了。杰森一把揪起小奶包抱在怀里，豪气万千， “宝贝儿，你放心，小白死人都能救，别说活生生的人了，你担心个屁，有空过来给我轰了黑手党的防护。”“刚是谁质疑白夜的医术的？”黑杰克对此表示疑惑， 杰森一掌过去，他敏捷闪开。小奶包被大高个子抱着，异常的纠结，踢了踢杰森，“放我下来。”“老子也想要这么个儿子，宁宁，你来当我儿子吧？老子垂涎你很久了。”杰森湛蓝色的眸迸发出澎湃的光芒，活似小奶包就是一块肥肉。众人，“......”白夜微笑说道，“杰森，你中文再让你妈教教，别老说长官不会用词语，你也好不到哪儿去。”“我和长官不是一个级别的好吧？”杰森很不满意有人把他和长官联系在一起，所谓官寇不一家，这是原则问题。小奶包挣扎一下，杰森放他下来，小奶包问道：“白夜......”“宁宁，等我给他们做过检查才能确定，你先别着急。”白夜说道，揉揉小奶包的头， “我保证会尽全力。”\\ \midrule
\model{Reference 1}        &  She called a maid to take Ye Chen and Cheng Anya downstairs to freshen up. Little Cutie really wanted to go with them, but he stayed behind as Bai Ye seemed ready to start fighting someone, so Little Cutie quickly grabbed his sleeve. ``Bai Ye, can you save my Mommy and Daddy?'' \textcolor{red}{The child's eyes were bright, like two black grapes embedded in his fair face, full of expectation. It seemed that if Bai Ye shook his head, the light in his eyes would dim. Jason grabbed Little Cutie and hugged him fiercely, feeling quite confident. ''Darling, don't worry. Little Bai can save even the dead, not to mention living people. Stop worrying about them and come help me take down the Mafia's line of defense when you have time.'' ``Who was questioning Bai Ye's medical skills just now?'' Black Jack expressed doubt about this. Jason threw a punch at him, but he agilely dodged it. \textcolor{blue}{Being carried by the tall man, Little Cutie felt extremely conflicted and kicked Jason. ''Put me down.'' ``I also want a son like you. Ningning, will you be my son? I've been coveting you for a long time.'' Jason's azure eyes burst with enthusiasm as if Little Cutie was a rare piece of jewelry. Everyone was speechless.... Bai Ye smiled and said, ''Jason, let your mother help you improve your Chinese. Don't always complain that the commander doesn't know how to use words. You're not much better.'' ``I'm not on the same level as the commander, okay?'' Jason was very dissatisfied that someone had compared him to the commander. It was a matter of principle that officials and bandits were not on the same side.} Little Cutie struggled a little, and Jason put him down. Little Cutie asked, ``Bai Ye....''} ``Ningning, don't worry. I'll check them up before making a conclusion.'' Bai Ye said, rubbing the head of Little Cutie, ''I promise I will do my best.''                                                             \\ \midrule
\model{gpt-4-1106-preview} &  She summoned the maids to take Ye Chen and Cheng Anya to wash up. Xiao Nai Bao wanted to follow, but he stayed here, as Bai Ye looked like he was about to beat someone up, so Xiao Nai Bao quickly grabbed his sleeve. Bai Ye, can you save my daddy and mommy? Xiao Nai Bao's eyes were bright, like two black grapes embedded in his tender face, full of hope, as if the light in his eyes would dim if Bai Ye shook his head. Jason grabbed Xiao Nai Bao and hugged him, full of pride, ``Baby, don't worry, Xiao Bai can even save the dead, let alone the living. Stop worrying about nothing, come help me bombard the mafia's defenses when you're free.'' Who was questioning Bai Ye's medical skills just now? Black Jack expressed doubt, and with a palm strike, Jason swiftly dodged. Xiao Nai Bao struggled a bit, and Jason put him down. Xiao Nai Bao asked, ``Bai Ye...'' Ningning, wait until I've examined them before I can be sure, don't be anxious, Bai Ye said, ruffling Xiao Nai Bao's head, ``I promise to do my best.''                                  \\ \midrule
\ours                      &  At her behest, maids ushered Ye Chen and Cheng An Ya away for a bath. Xiao Naibao, torn between following and staying, ultimately remained as Bai Ye appeared poised for a scuffle. Tugging at Bai Ye's sleeve, Xiao Naibao looked up at him with hopeful eyes. ``Bai Ye, is there a way to cure my daddy and mommy?'' ``Ning Ning, let me conduct a thorough examination first. Stay calm,'' Bai Ye soothed, patting the boy's head. ``I'll do everything in my power.''     \\ \bottomrule
\end{tabular}
\caption{
Case study for undertranslation.
The text highlighted in \textcolor{red}{red} and \textcolor{blue}{blue} indicates the content that is omitted by \ours.
The text highlighted in \textcolor{red}{red} represents the content that is omitted by \model{gpt-4-1106-preview}.
}
\label{tab:case_omission}
\end{table*}

\end{CJK*}
\paragraph{Undertranslation}
Our \ours is generally preferred over both \model{Reference 1} and \model{gpt-4-1106-preview} according to evaluations by human judges and large language models (LLMs) (\autoref{fig:mhp_eval} and \autoref{fig:blp_eval}). However, despite its higher preference, the translations produced by \model{\ours} are not without flaws. A detailed analysis of the translated chapters, when divided into smaller segments, reveals that both \model{gpt-4-1106-preview} and \model{\ours} exhibit significant issues with undertranslation, as illustrated in \autoref{tab:case_omission}. While these undertranslations do not seem to impact the overall development of the story plot, they could potentially influence other critical aspects of the narrative. For example, missing content could diminish the depth of character development or alter the intended emotional impact of the text. Such undertranslations, therefore, raise concerns about the completeness and fidelity of the translation in preserving the nuanced expressions and thematic elements of the original texts.

\begin{table*}[t]
\centering
\tiny
\begin{tabular}{@{}lp{0.8\textwidth}@{}}
\toprule
Translator A & \textcolor{red}{\ours}'s translation style is similar to that of a novel, with sophisticated wording and personal flair. Despite some omissions, it makes the text more concise and effectively conveys the original text's mood and meaning. \textcolor{red}{\model{Reference 1}} and \textcolor{red}{\model{gpt-4-1106-preview}}'s translations are more conventional, adhering strictly to the original text word for word. However, \textcolor{red}{\model{gpt-4-1106-preview}}'s translation is more grammatically precise than \textcolor{red}{\model{Reference 1}}'s, and its wording is slightly better, making its translation aesthetically superior to \textcolor{red}{\model{Reference 1}}'s but still not reaching the literary expressiveness of \textcolor{red}{\ours}. From their translation habits, \textcolor{red}{\ours} appears to have a solid foundation in English, \textcolor{red}{\model{Reference 1}} seems to rely on machine translation, and \textcolor{red}{\model{gpt-4-1106-preview}} behaves like a standard, rule-abiding translator. \\ \midrule
Translator B & \textcolor{red}{\ours}'s translation breaks away from the constraints of the original language, using the language freely with ample additions and expansions, and the choice of vocabulary also demonstrates a deeper understanding of the language. \textcolor{red}{\model{Reference 1}} remains faithful to the original text, translating directly and succinctly without adding personal interpretations. \textcolor{red}{\model{gpt-4-1106-preview}}'s translation style is similar to \textcolor{red}{\model{Reference 1}}'s, both strictly adhering to the original without much personal interpretation or embellishment. Overall, \textcolor{red}{\ours}'s translation shows the greatest depth and sophistication, followed by \textcolor{red}{\model{Reference 1}}, while \textcolor{red}{\model{gpt-4-1106-preview}} performs most ordinarily among the three. \\ \bottomrule                                                                                             
\end{tabular}
\caption{
Comments from two experienced professional translators on the translations from \ours, \model{Reference 1}, and \model{gpt-4-1106-preview}.
We present both the original text and the anonymized translations to two experienced professional translators. 
The original comments are written in Chinese, and we make adaptations while preserving their original meaning.
We replace the anonymized system names with the actual system names to improve readability.
The translation systems are highlighted in \textcolor{red}{red}.
}
\label{tab:case_comments}
\end{table*}

\paragraph{Comments from Professional Translators}
We anonymize the translations from \ours, \model{Reference 1}, and \model{gpt-4-1106-preview} for a randomly selected chapter and present both the original text and the translations to two experienced professional translators. We request that they assess and rank the quality of each translation and provide their comments on the translations. As shown in \autoref{tab:case_comments}, both Translator A's and Translator B's comments highlight the novel-like, expressive translation style of \ours, which uses sophisticated language, though it sometimes omits parts of the original text. \model{Reference 1}, and \model{gpt-4-1106-preview} stick closer to the original text. Overall, \ours's translations are viewed as the most expressive and engaging, \model{Reference 1}'s as straightforward, and \model{gpt-4-1106-preview}'s as the most traditional. These comments confirm that \ours is capable of producing more expressive and engaging translations, compared to \model{Reference 1} and \model{gpt-4-1106-preview}.


\begin{CJK*}{UTF8}{gbsn}

\begin{figure}[t]
    \centering
    \footnotesize
    \begin{Verbatim}[frame=single, fontsize=\tiny, breaklines=true, breakanywhere=true, commandchars=\\\{\}]
# Translation Guidelines

## Glossary
罗德: Rhode
虚空之龙: Void Dragon
星月佣兵团: Star Moon Mercenary Corps
[TRUNCATED]

## Book Summary
The book centers on Rhode Alante, initially a Summoner Swordsman in the game 'Dragon Soul Continent,' [TRUNCATED]

## Tone
The tone of the book is adventurous and immersive with elements of fantasy and suspense. [TRUNCATED]

## Style
The book is a gripping blend of fantasy and litRPG, characterized by its immersive world-building, dynamic combat scenes, and a clear progression system. [TRUNCATED]

## Target Audience
The target audience for this book includes young adults and adults who enjoy fantasy and adventure genres, particularly those who are fans of MMORPG [TRUNCATED]

# Chapter Text
序章 传奇落幕
乌云笼罩着天空，昏暗无光的地面上四周都是一片狼籍。
[TRUNCATED]

# Instruction
Translate the chapter text from Chinese into English Ensure that your translation closely adheres to the provided translation guidelines, including the glossary, book summary, tone, style, and target audience, for consistency and accuracy. Remember to maintain the original meaning and tone as much as possible while making the translation understandable in English.

    \end{Verbatim}
    \caption{An example prompt for the Translator, including the translation guidelines, the chapter text in the source language, and the instruction.}
    \label{fig:prompt_translation}
\end{figure}

\end{CJK*}


\begin{CJK*}{UTF8}{gbsn}

\begin{figure}[t]
    \centering
    \footnotesize
    \begin{Verbatim}[frame=single, fontsize=\tiny, breaklines=true, breakanywhere=true, commandchars=\\\{\}]
# Translation Guidelines

## Glossary
罗德: Rhode
虚空之龙: Void Dragon
星月佣兵团: Star Moon Mercenary Corps
[TRUNCATED]

## Book Summary
The book centers on Rhode Alante, initially a Summoner Swordsman in the game 'Dragon Soul Continent,' [TRUNCATED]

## Tone
The tone of the book is adventurous and immersive with elements of fantasy and suspense. [TRUNCATED]

## Style
The book is a gripping blend of fantasy and litRPG, characterized by its immersive world-building, dynamic combat scenes, and a clear progression system. [TRUNCATED]

## Target Audience
The target audience for this book includes young adults and adults who enjoy fantasy and adventure genres, particularly those who are fans of MMORPG [TRUNCATED]

# Chapter Text
序章 传奇落幕
乌云笼罩着天空，昏暗无光的地面上四周都是一片狼籍。
[TRUNCATED]

# Chapter Translation
Prologue, the end of the legend
Dark clouds hung over the sky, and there was a mess all around on the dim ground.
[TRUNCATED]

# Instruction
Guided by our translation guidelines, including glossary, book summary, tone, style, and target audience, localize the chapter translation for English context. You MUST maintain all the details and the orginal writing style of the chapter text.

    \end{Verbatim}
    \caption{An example prompt for the Localization Specialist, including the translation guidelines, the chapter text in the source language, the chapter translation in the target language, and the instruction.}
    \label{fig:prompt_localization}
\end{figure}

\end{CJK*}


\begin{CJK*}{UTF8}{gbsn}

\begin{figure}[t]
    \centering
    \footnotesize
    \begin{Verbatim}[frame=single, fontsize=\tiny, breaklines=true, breakanywhere=true, commandchars=\\\{\}]
# Translation Guidelines

## Glossary
罗德: Rhode
虚空之龙: Void Dragon
星月佣兵团: Star Moon Mercenary Corps
[TRUNCATED]

## Book Summary
The book centers on Rhode Alante, initially a Summoner Swordsman in the game 'Dragon Soul Continent,' [TRUNCATED]

## Tone
The tone of the book is adventurous and immersive with elements of fantasy and suspense. [TRUNCATED]

## Style
The book is a gripping blend of fantasy and litRPG, characterized by its immersive world-building, dynamic combat scenes, and a clear progression system. [TRUNCATED]

## Target Audience
The target audience for this book includes young adults and adults who enjoy fantasy and adventure genres, particularly those who are fans of MMORPG [TRUNCATED]

# Chapter Text
序章 传奇落幕
乌云笼罩着天空，昏暗无光的地面上四周都是一片狼籍。
[TRUNCATED]

# Chapter Translation
Prologue End of the Legend
Black clouds covered the sky. The ground was in darkness, with everything around in chaos.
[TRUNCATED]

# Instruction
Guided by our translation guidelines, including the glossary, book summary, tone, style, and target audience, proofread the chapter translation.

    \end{Verbatim}
    \caption{An example prompt for the Proofreader, including the translation guidelines, the chapter text in the source language, the chapter translation in the target language, and the instruction.}
    \label{fig:prompt_proofreading}
\end{figure}

\end{CJK*}

\revised{
\section{Prompts}
\label{sec:prompts_app}
In this section, we present the prompts used for translation, localization, and proofreading in \autoref{fig:prompt_translation}, \autoref{fig:prompt_localization}, and \autoref{fig:prompt_proofreading}, respectively.
}

\end{document}